\documentclass[11pt,a4paper]{article}
\usepackage[hyperref]{acl2020}
\usepackage{times}
\usepackage{latexsym}

\usepackage{microtype}

\usepackage{multirow}
\usepackage{adjustbox}
\usepackage{amssymb,amsmath}
\usepackage{algorithm}      
\usepackage{algpseudocode}
\usepackage{bm}            
\usepackage{subfigure}
\usepackage{amsfonts}
\usepackage{makecell}

\aclfinalcopy
\title{Depth-Adaptive Graph Recurrent Network for Text Classification}

\author{
  Yijin Liu\textsuperscript{1}\thanks{\ \ This work was done when Yijin Liu was interning at Pattern Recognition Center, WeChat AI, Tencent Inc, China} \ ,
  Fandong Meng\textsuperscript{2}, 
  Jie Zhou\textsuperscript{2},
  Yufeng Chen\textsuperscript{1} 
  and Jinan Xu\textsuperscript{1}\thanks{ \ \ Jinan Xu is the corresponding author of the paper.} \\
  \textsuperscript{1}Beijing Jiaotong University, China \\
  \textsuperscript{2}Pattern Recognition Center, WeChat AI, Tencent Inc, China \\
  \texttt{adaxry@gmail.com} \\
  \texttt{\{fandongmeng, withtomzhou\}@tencent.com} \\
  \texttt{\{chenyf, jaxu\}@bjtu.edu.cn} \\
}

\date{}

\begin{document}
\maketitle

\begin{abstract}
The Sentence-State LSTM (S-LSTM) \cite{slstm_2018} is a powerful and high efficient graph recurrent network, which views words as nodes and performs layer-wise recurrent steps between them simultaneously. Despite its successes on text representations, the S-LSTM still suffers from two drawbacks. 
Firstly, given a sentence, certain words are usually more ambiguous than others, and thus more computation steps need to be taken for these difficult words and vice versa. However, the S-LSTM takes fixed computation steps for all words, irrespective of their hardness. 
The secondary one comes from the lack of sequential information ({\em e.g.}, word order) that is inherently important for natural language. In this paper, we try to address these issues and propose a depth-adaptive mechanism for the S-LSTM, which allows the model to learn how many computational steps to conduct for different words as required.
In addition, we integrate an extra RNN layer to inject sequential information,
which also serves as an input feature for the  decision of adaptive depths.
Results on the classic text classification task (24 datasets in various sizes and domains) show that our model brings significant improvements against the conventional S-LSTM and other high-performance models ({\em e.g.,} the Transformer), meanwhile achieving a good accuracy-speed trade off.
\end{abstract}

\begin{figure}[t!]
\begin{center}
     \scalebox{0.43}{
       \includegraphics[width=1\textwidth]{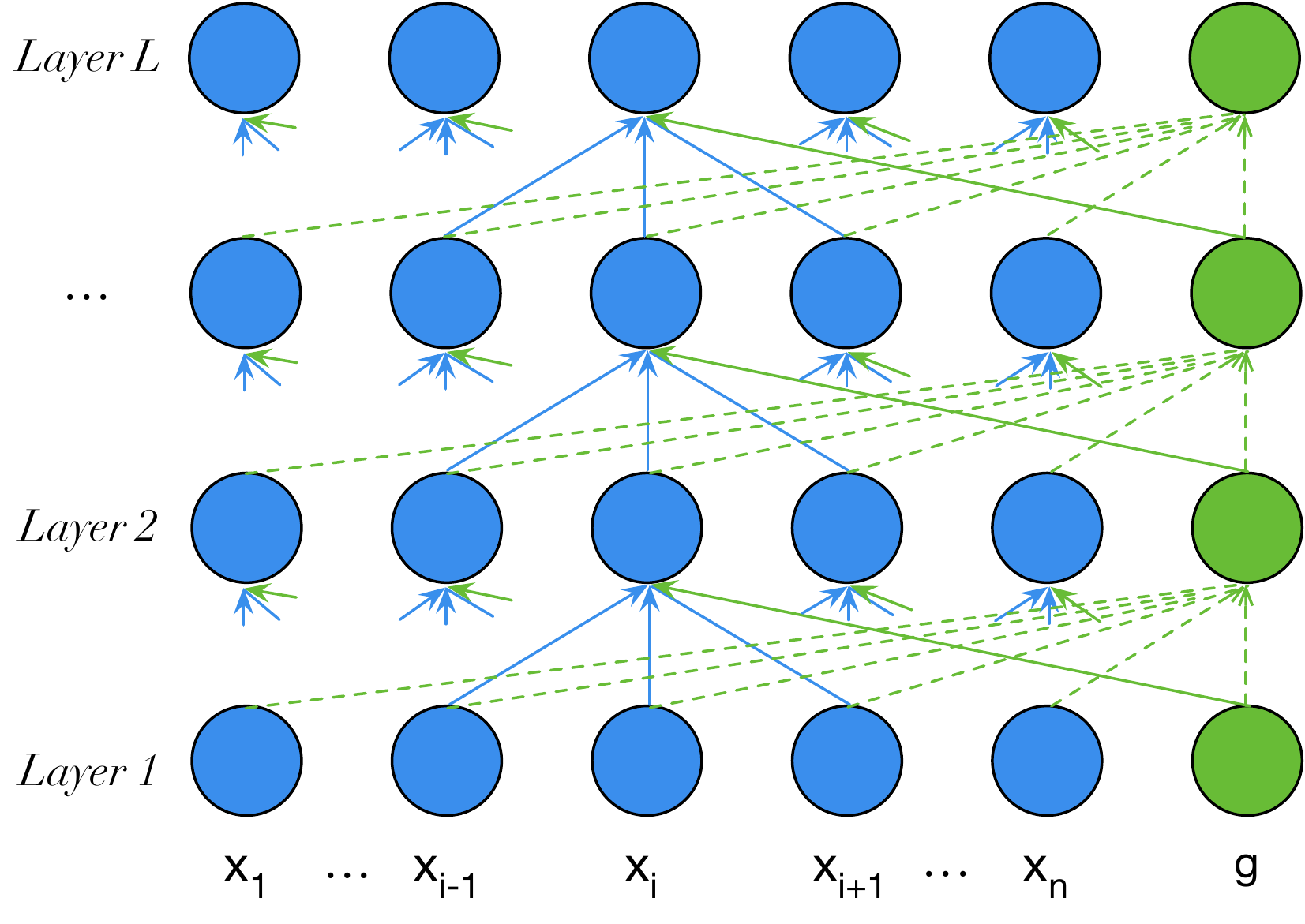}
      } \vspace{-5pt}
      \caption{Process of recurrent state transition in the S-LSTM. Given an input sentence with $n$ words, in each layer, the word $x_i$ takes information from its predecessor $x_{i-1}$, successor $x_{i+1}$, the global node $g$ and itself to update its hidden state (solid lines). Meanwhile, the global node $g$ takes all local states including itself from the previous layer as context vectors to update global state (dashed line). Both update operations take place simultaneously, and layer-wise parameters are shared.} \vspace{-20pt}
      \label{slstm}
 \end{center}
\end{figure}

\section{Introduction}

Recent advances of graph recurrent network (GRN) have shown impressive performance in many tasks, including sequence modeling \cite{slstm_2018}, sentence ordering \cite{sentence_ordering_2019}, machine translation \cite{slstm_mt_2018, densely_slstm_2019}, and spoken language understanding \cite{cmnet_2019}. Among these neural networks, the representative S-LSTM has drawn great attention for its high efficiency and
strong representation capabilities.
More specifically, it views a sentence as a graph of word nodes, and performs layer-wise recurrent steps between words simultaneously, rather than incrementally reading a sequence of words in a sequential manner ({\em e.g.,} RNN).  
Besides the local state for each individual word, the S-LSTM preserves a shared global state for the overall sentence. Both local and global states get enriched incrementally by exchanging information between each other. A visual process of recurrent state transition in the S-LSTM is shown in Figure \ref{slstm}. 

In spite of  its successes, there still exist several limitations in the S-LSTM.
For example, given a sentence, certain words are usually more ambiguous than others. Considering this example `The film was awesome ...', whether `awesome' means thrilling or excellent is a confusion, thus more contexts should be taken and more layers of abstraction are necessary to refine feature representations. 
One possible solution is to simply train very deep networks over all word positions, irrespective of their hardness, that is exactly what the conventional S-LSTM does. However, in terms of both computational efficiency and ease of learning, it is preferable to allow model itself to `ponder' and `determine' how many steps of computation to take at each position~\cite{ACT_2016, UT_2019}.

In this paper, we focus on addressing the above issue in the S-LSTM, and propose a depth-adaptive mechanism that enables the model to adapt depths as required. 
Specifically, at each word position, the  \textit{executed} depth is firstly determined by a
specific classifier with corresponding input features, and proceeds to iteratively refine representation until reaching its own \textit{executed} depth. We also investigate different strategies to obtain the depth distribution, and further endow the model with depth-specific vision through  a novel depth embedding.

Additionally, the parallel nature of the S-LSTM makes it inherently lack in modeling sequential information ({\em e.g.,} word order), which has been shown a highly useful complement to the no-recurrent models
\cite{best_2018,r_transformer_2019}.
We investigate different ways to integrate RNN's inductive bias into our model. 
Empirically, our experiments indicate this inductive bias is of great matter for text representations.
Meanwhile, the informative representations emitted by the RNN are served as input features to calculate the \textit{executed} depth in our depth-adaptive mechanism.

To evaluate the effectiveness and efficiency of our proposed model, we conduct extensive experiments on the text classification task with 24 datasets in various sizes and domains. Results on all datasets show that our model significantly outperforms the conventional S-LSTM, and other strong baselines ({\em e.g.,} stacked Bi-LSTM, the Transformer) while achieving a good accuracy-speed trade off. 
Additionally, our model achieves state-of-the-art performance on 16 out of total 24 datasets.

Our main contributions are as follows\footnote{Code is available at: https://github.com/Adaxry/Depth-Adaptive-GRN}:
\begin{itemize}
\item We are the first to investigate a depth-adaptive mechanism on graph recurrent network, and significantly boost the performance of the representative S-LSTM model.
\item We empirically verify the effectiveness and necessity of recurrent inductive bias for the S-LSTM.
\item Our model consistently outperforms strong baseline models and achieves state-of-the-art performance on 16 out of total 24 datasets.
\item We conduct thorough analysis to offer more insights and elucidate the properties of our model. Consequently, our depth-adaptive model achieves a good accuracy-speed trade off when compared with full-depth models.
\end{itemize}

\begin{figure*}[t!]
\begin{center}
     \scalebox{0.88}{
       \includegraphics[width=1\textwidth]{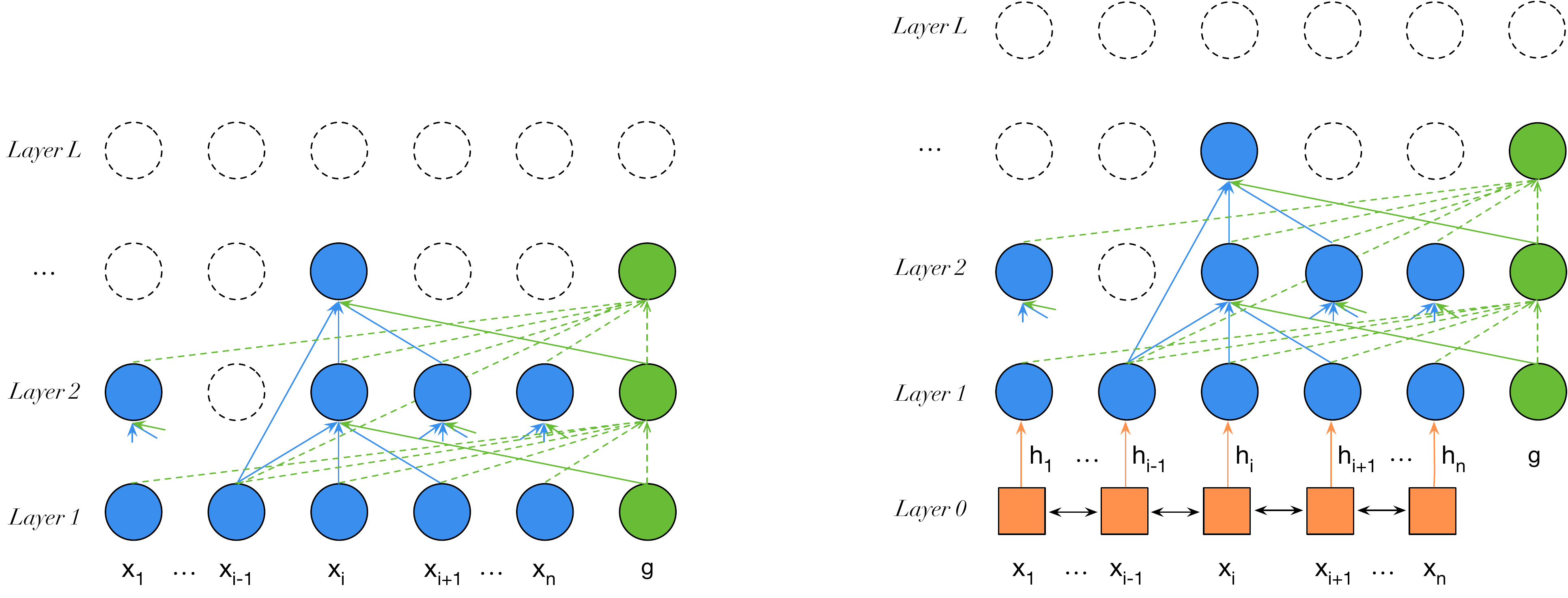}
      }\vspace{-8pt}
      \caption{Overview of our proposed model (left part), whose \textit{executed} depth is varying at different word positions. The dashed nodes indicate that their sates are directly copied from lower layers without computation.
      In addition, we introduce Bi-RNN (orange squares in the right part) at the bottom layer for two usages: (1) providing sequential information for upper modules, and (2) serving as input features for the calculation of \textit{executed} depths.} 
      \vspace{-12pt} \label{slstm_adaptive} 
 \end{center} 
\end{figure*}

\section{Background}
Formally, in the $l$-{th} layer of the S-LSTM, hidden states and cell states can be denoted by:
\begin{equation}
\setlength{\abovedisplayskip}{5pt}
\setlength{\belowdisplayskip}{5pt}
    \begin{split}
    \boldsymbol{H}^{l} &= \{\boldsymbol{h}_{1}^{l}, \boldsymbol{h}_{2}^{l}, \dots,
    \boldsymbol{h}_{n}^{l}, \boldsymbol{g}^{l}\} \\
    \boldsymbol{C}^{l} &= \{\boldsymbol{c}_{1}^{l}, \boldsymbol{c}_{2}^{l}, \dots,
    \boldsymbol{c}_{n}^{l}, \boldsymbol{c}_{g}^{l}\} 
    \end{split}
\end{equation}
where $\boldsymbol{h}_{i}^{l}$ ($i \in [1, n]$) is the hidden state for the $i$-th word, and $\boldsymbol{g}^{l}$ is the hidden state for the entire sentence. Similarly for cell states $\boldsymbol{C}^{l}$. Note that $n$ is the number of words for a sentence, and the $0$-{th} and ($n$+$1$)-{th} words are padding signals.

As shown in Figure \ref{slstm}, the states transition from $\boldsymbol{H}^{l-1}$ to $\boldsymbol{H}^{l}$ consists of two parts: (1)  word-level transition from $\boldsymbol{h}_{i}^{l-1}$ to  $\boldsymbol{h}_{i}^{l}$; (2) sentence-level transition from $\boldsymbol{g}^{l-1}$ to $\boldsymbol{g}^{l}$. The former process is computed as follows:

\begin{equation}
\setlength{\abovedisplayskip}{3pt}
\setlength{\belowdisplayskip}{3pt}
\begin{split}
\boldsymbol{\xi}_{i}^{l} &=\left[\boldsymbol{h}_{i-1}^{l-1}, \boldsymbol{h}_{i}^{l-1}, \boldsymbol{h}_{i+1}^{l-1}\right] \\
\hat{\boldsymbol{l}}_{i}^{l}
&=\sigma\left(\boldsymbol{W}_{l} \boldsymbol{\xi}_{i}^{l}+\boldsymbol{U}_{l} \boldsymbol{x}_{i}+\boldsymbol{V}_{l} \boldsymbol{g}^{l-1}+\boldsymbol{b}_{l}\right) \\
\hat{\boldsymbol{r}}_{i}^{l} &=\sigma\left(\boldsymbol{W}_{r} \boldsymbol{\xi}_{i}^{l}+\boldsymbol{U}_{r} \boldsymbol{x}_{i}+\boldsymbol{V}_{r} \boldsymbol{g}^{l-1}+\boldsymbol{b}_{r}\right) \\
\hat{\boldsymbol{i}}_{i}^{l}
&=\sigma\left(\boldsymbol{W}_{i} \boldsymbol{\xi}_{i}^{l}+\boldsymbol{U}_{i} \boldsymbol{x}_{i}+\boldsymbol{V}_{i} \boldsymbol{g}^{l-1}+\boldsymbol{b}_{i}\right) \\
\hat{\boldsymbol{f}}_{i}^{l} &=\sigma\left(\boldsymbol{W}_{f} \boldsymbol{\xi}_{i}^{l}+\boldsymbol{U}_{f} \boldsymbol{x}_{i}+\boldsymbol{V}_{f} \boldsymbol{g}^{l-1}+\boldsymbol{b}_{f}\right) \\ \hat{\boldsymbol{s}}_{i}^{l} &=\sigma\left(\boldsymbol{W}_{s} \boldsymbol{\xi}_{i}^{l}+\boldsymbol{U}_{s} \boldsymbol{x}_{i}+\boldsymbol{V}_{s} \boldsymbol{g}^{l-1}+\boldsymbol{b}_{s}\right) \\ \boldsymbol{o}_{i}^{l} &=\sigma\left(\boldsymbol{W}_{o} \boldsymbol{\xi}_{i}^{l}+\boldsymbol{U}_{o} \boldsymbol{x}_{i}+\boldsymbol{V}_{o} \boldsymbol{g}^{l-1}+\boldsymbol{b}_{o}\right) \\ \boldsymbol{u}_{i}^{l} &=\tanh \left(\boldsymbol{W}_{u} \boldsymbol{\xi}_{i}^{l}+\boldsymbol{U}_{u} \boldsymbol{x}_{i}+\boldsymbol{V}_{u} \boldsymbol{g}^{l-1}+\boldsymbol{b}_{u}\right) \\ 
\boldsymbol{i}_{i}^{l}, &\boldsymbol{l}_{i}^{l}, \boldsymbol{r}_{i}^{l}, \boldsymbol{f}_{i}^{l}, \boldsymbol{s}_{i}^{l}=\operatorname{softmax}\left(\hat{\boldsymbol{i}}_{i}^{l}, \hat{\boldsymbol{l}}_{i}^{l}, \hat{\boldsymbol{r}}_{i}^{l}, \hat{\boldsymbol{f}}_{i}^{l}, \hat{\boldsymbol{s}}_{i}^{l}\right) \\
\boldsymbol{c}_{i}^{l}
&=\boldsymbol{l}_{i}^{l} \odot \boldsymbol{c}_{i-1}^{l-1} +\boldsymbol{f}_{i}^{l} \odot \boldsymbol{c}_{i}^{l-1}+\boldsymbol{r}_{i}^{l} \odot \boldsymbol{c}_{i+1}^{l-1} \\
& \ \ \ \ \ \ \ \ \ \ \ \ \ \ \ \ \ \ \ \ \ +\boldsymbol{s}_{i}^{l} \odot \boldsymbol{c}_{g}^{l-1}+\boldsymbol{i}_{i}^{l} \odot \boldsymbol{u}_{i}^{l} \\
\boldsymbol{h}_{i}^{l}
&=\boldsymbol{o}_{l}^{i} \odot \tanh \left(\boldsymbol{c}_{i}^{l}\right)
\end{split}
\label{back_local_update}
\end{equation}
where $\boldsymbol{\xi}_{i}^{l}$ is the concatenation of hidden states in a window, and $\boldsymbol{l}_{i}^{l}$,  $\boldsymbol{r}_{i}^{l}$, $\boldsymbol{f}_{i}^{l}$ and $\boldsymbol{s}_{i}^{l}$ are forget gates for left $\boldsymbol{c}_{i-1}^{l-1}$, right  $\boldsymbol{c}_{i+1}^{l-1}$, corresponding $\boldsymbol{c}_{i}^{l-1}$ and sentence-level cell state $\boldsymbol{c}_{g}^{l-1}$. $\boldsymbol{i}_{i}^{l}$ and $\boldsymbol{o}_{i}^{l}$ are input and output gates. The value of all gates are normalised such that they sum to $1$. $\boldsymbol{W}_{*}$, $\boldsymbol{U}_{*}$, $\boldsymbol{V}_{*}$ and $\boldsymbol{b}_{*}$ ($* \in \{l, r, f, s, i, o, u\}$) are model parameters.

Then the state transition of sentence-level $\boldsymbol{g}^{l}$ is computed as follows:
\begin{equation}
\setlength{\abovedisplayskip}{5pt}
\setlength{\belowdisplayskip}{5pt}
    \begin{split}
        \overline{\boldsymbol{h}} &=\operatorname{avg}\left(\boldsymbol{h}_{1}^{l-1}, \boldsymbol{h}_{2}^{l-1}, \ldots, \boldsymbol{h}_{n}^{l-1}\right) \\
        \hat{\boldsymbol{f}}_{g}^{l} &=\sigma\left(\boldsymbol{W}_{g} \boldsymbol{g}^{l-1}+\boldsymbol{U}_{g} \overline{\boldsymbol{h}}+\boldsymbol{b}_{g}\right) \\
        \hat{\boldsymbol{f}}_{i}^{l} &=\sigma\left(\boldsymbol{W}_{f} \boldsymbol{g}^{l-1}+\boldsymbol{U}_{f} \boldsymbol{h}_{i}^{l-1}+\boldsymbol{b}_{f}\right) \\
        \boldsymbol{o}^{l} &=\sigma\left(\boldsymbol{W}_{o} \boldsymbol{g}^{l-1}+\boldsymbol{U}_{o} \overline{\boldsymbol{h}}+\boldsymbol{b}_{o}\right) \\
        \boldsymbol{f}_{1}^{l} &, \ldots, \boldsymbol{f}_{n}^{l}, \boldsymbol{f}_{g}^{l}=\operatorname{softmax}\left(\hat{\boldsymbol{f}}_{1}^{l}, \ldots, \hat{\boldsymbol{f}}_{n}^{l}, \hat{\boldsymbol{f}}_{g}^{l}\right) \\ 
        \boldsymbol{c}_{g}^{l} &=\boldsymbol{f}_{g}^{l} \odot \boldsymbol{c}_{g}^{l-1}+\sum_{i} \boldsymbol{f}_{i}^{l} \odot \boldsymbol{c}_{i}^{l-1} \\
        \boldsymbol{g}^{l} &=\boldsymbol{o}^{l} \odot \tanh \left(\boldsymbol{c}_{g}^{l}\right)
    \end{split} \label{back_global_update}
\end{equation}    
where $\boldsymbol{f}_{1}^{l}, \dots, \boldsymbol{f}_{n}^{l}, \boldsymbol{f}_{g}^{l}$ are normalised gates for controlling $\boldsymbol{c}_{1}^{l-1}, \dots, \boldsymbol{c}_{n}^{l-1}, \boldsymbol{c}_{g}^{l-1}$, respectively. $\boldsymbol{o}^{l}$ is an output gate, and $\boldsymbol{W}_{*}$, $\boldsymbol{U}_{*}$ and $\boldsymbol{b}_{*}$ ($* \in \{f, g, o\}$) are model parameters.

\section{Model}
As the overview shown in Figure \ref{slstm_adaptive}, our model conducts dynamic steps across different positions, which is more sparse than the conventional S-LSTM drawn in Figure \ref{slstm}. We then proceed to more details of our model in the following sections.

\subsection{Token Representation} 
Given an input sentence $S = \{x_1, x_2, \cdots, x_n \}$ with $n$ words, we firstly obtain word embeddings $\boldsymbol{x}^{glove}$ from the lookup table initialized by Glove\footnote{https://nlp.stanford.edu/projects/glove/}. Then we train character-level word embeddings from scratch by Convolutional Neural Network (CNN) \cite{first_CNN_2014}. The glove and character-level embeddings are concatenated to form the final token representations $\boldsymbol{X} = \{\boldsymbol{x}_{1}, \dots, \boldsymbol{x}_{n} \}$:
\begin{equation}
\setlength{\abovedisplayskip}{3pt}
\setlength{\belowdisplayskip}{3pt}
    \begin{split}
        \boldsymbol{x}_{i} = [\boldsymbol{x}_{i}^{glove};\boldsymbol{x}_{i}^{char}] 
    \end{split} \label{token_representation}
\end{equation}

\subsection{Sequential Module}
As mentioned above, the conventional S-LSTM identically treats all positions, and fails to utilize the order of an input sequence. 
We simply build one layer Bi-LSTMs\footnote{We also experiments with using learnable or sinusoidal position embedding \cite{pos_emb_2017}, and choose to a simple RNN layer considering the trade off between accuracy and efficiency. More details in Section \ref{ablation_section}.} upon the word embedding layer to inject sequential information (right part in Figure \ref{slstm_adaptive}), which is computed as follows:
\begin{equation}
\setlength{\abovedisplayskip}{6pt}
\setlength{\belowdisplayskip}{6pt}
    \begin{split}
    & \overrightarrow{\boldsymbol{h}_i} = \overrightarrow{\boldsymbol{LSTM}}(\boldsymbol{x}_i, \overrightarrow{\boldsymbol{h}}_{i-1}; \boldsymbol{\overrightarrow{\theta}}) \\
    & \overleftarrow{\boldsymbol{h}_i} = \overleftarrow{\boldsymbol{LSTM}}(\boldsymbol{x}_i, \overleftarrow{\boldsymbol{h}}_{i+1}; \boldsymbol{\overleftarrow{\theta}}) \\
    & \boldsymbol{h}_i = [\overrightarrow{\boldsymbol{h}_i}; \overleftarrow{\boldsymbol{h}_i}] \\
    \end{split}
    \label{bilsmt_ouput}
\end{equation}
where $\boldsymbol{\overrightarrow{\theta}}$ and $\boldsymbol{\overleftarrow{\theta}}$ are parameter sets of Bi-LSTMs. 
The output hidden states $\boldsymbol{H} = \{\boldsymbol{h}_{1}, \boldsymbol{h}_{2}, \dots, \boldsymbol{h}_{n} \}$ also serve as input features for the following depth-adaptive mechanism.

\subsection{Depth-Adaptive Mechanism}
In this section, we describe how to dynamically calculate the depth for each word, and use it to control the state transition process in our model. Specifically, for the $i$-{th} word ($i \in[1, n]$) in a sentence,  
its hidden state $\boldsymbol{h}_{i} \in \mathbb{R}^{d_{model}} $ is fed to a fully connected feed-forward network \cite{Transformer_2017} to calculate logits value $\boldsymbol{l}_{i}$ of depth distribution:
\begin{equation}
\setlength{\abovedisplayskip}{5pt}
\setlength{\belowdisplayskip}{5pt}
    \begin{split}
        \boldsymbol{l}_{i} = \max \left(0, \boldsymbol{h}_{i} \boldsymbol{W}_{1} + \boldsymbol{b}_{1}\right) \boldsymbol{W}_{2} + 
        \boldsymbol{b}_{2}
    \end{split}\label{equ_logits}
\end{equation}
where $\boldsymbol{W}_{1} \in \mathbb{R}^{d_{model} \times d_{inner}}$ is a matrix that maps $\boldsymbol{h}_{i}$ into an inner vector, and $\boldsymbol{W}_{2} \in \mathbb{R}^{d_{inner} \times L}$ is a matrix that maps the inner vector into a $L$-dimensional vector, and $L$ denotes a predefined number of maximum layer.
Then the probability ${p}_{i}^{j}$ of the $j$-{th} depth is computed by $\operatorname{softmax}$:
\begin{equation}
\setlength{\abovedisplayskip}{5pt}
\setlength{\belowdisplayskip}{5pt}
p_{i}^{j} =
\frac{e^{l_{i}^{j}}}{\sum_{k=1}^{L} e^{l_{i}^{k}}} \quad \text { for } j=1, \ldots, L
\label{equ_depth_prob}
\end{equation}

In particular, we consider three ways to select the depth $d_i$ from the probability ${p}_{i}^{j}$.
\paragraph{Hard Selection:} 
The most direct way is to choose the number with the highest probability from the depth distribution drawn by Eq. (\ref{equ_depth_prob}):
\begin{equation}
\setlength{\abovedisplayskip}{5pt}
\setlength{\belowdisplayskip}{5pt}
    d_{i} = \operatorname{argmax}(\boldsymbol{p}_{i}) \label{equ_deph_argmax}
\end{equation}

\paragraph{Soft Selection:}  A smoother version of selection is to sum up each depth weighted by the corresponding probability. We floor the value considering the discrete nature of the depth distribution by
\begin{equation}
\setlength{\abovedisplayskip}{5pt}
\setlength{\belowdisplayskip}{5pt}
    d_{i} = \left \lfloor \sum_{j=0}^{L-1}{ j \times p_{i}^{j}} \right \rfloor
    \label{equ_deph_soft_argmax} 
\end{equation}

\paragraph{Gumbel-Max Selection:}  
For better simulating the discrete distribution and more robust depth selection, we use Gumbel-Max \cite{gumbel_1954,gumbel_2014}, which provides an efficient and robust way to sample from a categorical distribution. Specifically, we add independent Gumbel perturbation $\boldsymbol{\eta}_{i}$ to each logit $\boldsymbol{l}_{i}$ drawn by Eq. (\ref{equ_logits}):
\begin{equation}
\setlength{\abovedisplayskip}{5pt}
\setlength{\belowdisplayskip}{5pt}
\begin{split}
\boldsymbol{\eta}_{i} &=-\log (-\log \boldsymbol{u}_{i}) \\ 
\widetilde{\boldsymbol{l}}_{i} &= (\boldsymbol{l}_{i} + \boldsymbol{\eta}_{i}) / \tau
\end{split}
\end{equation}
where $\boldsymbol{\eta}_{i}$ is computed from a uniform random variable $u \sim \mathcal{U}(0,1)$, and $\tau$ is temperature. As $\tau \rightarrow 0$, samples from the perturbed distribution $\widetilde{\boldsymbol{l}}_{i}$ become one-hot, and become uniform as $\tau \rightarrow \infty$. After that, the exact number of depth $d_i$ is calculated by modifying the Eq. (\ref{equ_depth_prob}) to:
\begin{equation}
\setlength{\abovedisplayskip}{5pt}
\setlength{\belowdisplayskip}{5pt}
    p_{i}^{j} = \frac{e^{\widetilde{l}_{i}^{j}}}{\sum_{k=1}^{L} e^{\widetilde{l}_{i}^{k}}} \quad \text { for } j=1, \ldots, L
   \label{equ_gumbel_softmax}
\end{equation}
Empirically, we set a tiny value to $\tau$, so the depth distribution calculated by Eq. (\ref{equ_gumbel_softmax}) is in the form of one-hot.
Note that Gumbel perturbations are merely used to select depths, and they would not affect the loss function for training.

\begin{table*}[t!]
\begin{center}
\scalebox{0.9}{
\begin{tabular}{l|c c c c c c c} 
\hline
\textbf{Dataset} & Classes & Type & \makecell{Average \ \ \\ \ \ Lenghts \ \ \  } & \makecell{\ \ Max \ \ \\ \ \ Lengths} &
\makecell{Train \ \ \\ \ \  Sample \ \ } & \makecell{Test \ \ \\ \ \ Sample \ \ } \\
\hline
TREC  \cite{trec_2002}      & 6        & Question   & 12   & 39       & 5,952   &  500    \\
AG’s News \cite{char_cnn_2015}  & 4        & Topic      & 44   & 221     & 120,000 &  7,600  \\
DBPedia  \cite{char_cnn_2015}     & 14       & Topic      & 67   & 3,841   & 560,000 &  70,000 \\
Subj \cite{subj_2004}        & 2        & Sentiment  & 26   & 122      & 10,000  &  CV   \\
MR \cite{MR_2005}         & 2        & Sentiment  & 23   & 61       & 10,622  &  CV   \\
Amazon-16 \cite{16_cls_data_17} *   & 2        & Sentiment  & 133  & 5,942   & 31,880  &  6,400 \\
IMDB \cite{IMDB_2011}        & 2        & Sentiment  & 230  & 2,472   & 25,000  &  25,000 \\  
Yelp Polarity \cite{char_cnn_2015} & 2        & Sentiment  & 177  & 2,066    & 560,000 &  38,000 \\ 
Yelp Full \cite{char_cnn_2015} & 5  & Sentiment  & 179  & 2,342    & 650,000 &  50,000 \\
\hline
\end{tabular}}
\end{center}  \vspace{-5pt}
\caption{Dataset statistics. `CV' means that there was no standard train/test split and thus 10-fold CV was used.
`$*$': There are 16 subsets with the same size in Amazon-16, named as Apparel, Baby, Books, Camera, DVD, Electronics, Health, IMDB, Kitchen, Magazines, MR, Music, Software, Sports, Toys and Video.
} \vspace{-8pt}
\label{data_statistics}
\end{table*}

After acquiring the depth number $d_{i}$ for each individual word,  additional efforts should be taken to connect the depth number $d_i$ with corresponding steps of computation.
Since our model has no access to explicit supervision for depth, in order to make our model learn such relevance, we must inject some depth-specific information into our model. To this end, we preserve a trainable depth embedding $\boldsymbol{x}^{depth}$ whose parameters are shared with the $\boldsymbol{W}_{2}$ in the above feed-forward network in Eq. (\ref{equ_logits}). 
We also sum a sinusoidal depth embedding with $\boldsymbol{x}^{depth}$ for the similar motivation with the Transformer \cite{Transformer_2017}:
\begin{equation}
\setlength{\abovedisplayskip}{6pt}
\setlength{\belowdisplayskip}{6pt}
\begin{split}
DE_{(d, 2j)} & =\sin (d / 10000^{2j / dim_{emb}}) \\
DE_{(d, 2j+1)} & =\cos (d / 10000^{2j/dim_{emb}}) \\
\end{split}  \label{pos_emb}  
\end{equation}
where $d$ is the depth, $dim_{emb}$ is the  the dimension of the depth embedding, and $j$ is index of $dim_{emb}$. As thus, the final token representation described by Eq. (\ref{token_representation}) is refined by:
\begin{equation}
\setlength{\abovedisplayskip}{5pt}
\setlength{\belowdisplayskip}{5pt}
    \begin{split}
        \boldsymbol{x}_{i} = [\boldsymbol{x}_{i}^{glove};\boldsymbol{x}_{i}^{char};\boldsymbol{x}_{i}^{depth}] 
    \end{split} \label{token_representation_refine}
\end{equation}

Then our model proceeds to perform dynamic state transition between words simultaneously.
More specifically, once a word $x_i$ reaches its own maximum layer $d_i$, it will stop state transition, and simply copy its state to the next layer until all words stop or the predefined maximum layer $L$ is reached. Formally, for the $i$-{th} word, its hidden state $\boldsymbol{h}_i$ is updated as follows:
\begin{equation}
\setlength{\abovedisplayskip}{5pt}
\setlength{\belowdisplayskip}{5pt}
\boldsymbol{h}_i^{l} =
\begin{cases}
\boldsymbol{h}_{i}^{l-1} \text{ \ \ \ \ \ \ \ \ \ \ \ \ \ \ \ \ \ \ \ \ \ \ \ \ \ \ \ \ \ \ \ \ \ \ \ \ \ \ \ \ \ \ \ \ \ \ if $ l > d_{i}$}  \\
S\textrm{-}LSTM(\boldsymbol{x}_{i}, \boldsymbol{h}_{i-1}^{l-1}, \boldsymbol{h}_{i}^{l-1}, \boldsymbol{h}_{i+1}^{l-1}, \boldsymbol{g}^{l-1})  \text{\ else} 
\end{cases}
\end{equation}
where $l \in [1, d_{max}]$ refers to the number of current layer, and $d_{max}$ is the maximum depth in current sentence. 
Specially, $\boldsymbol{h}_{i}^{0}$ is initialized by a linear transformation of the inner vector\footnote{Choosing the inner vector is for the gradient estimation when back propagation.} in Eq. (\ref{equ_logits}). $S\textrm{-}LSTM(\cdot)$ is the state transition function drawn by Eq. (\ref{back_local_update}). As the global state $\boldsymbol{g}$ is expected to encode the entire sentence, it conducts $d_{max}$ steps by default, which is drawn by Eq. (\ref{back_global_update}).

\subsection{Task-specific Settings}
After dynamic steps of computation among all nodes, we build task-specific models for the classification task. The output hidden states of the final layer $\boldsymbol{H}^{d_{max}}$ are firstly reduced by max and mean pooling. We then take the concatenations of these two reduced vectors and global states $\boldsymbol{g}^{d_{max}}$ to form the final feature vector $\boldsymbol{v}$. After the ReLU activation, $\boldsymbol{v}$ is fed to a softmax classification layer. Formally, the above-mentioned procedures are computed as follows:
\begin{equation}
\setlength{\abovedisplayskip}{6pt}
\setlength{\belowdisplayskip}{6pt}
    \begin{split}
        & \boldsymbol{v} = \operatorname{ReLU}([\mathop{\max}(\boldsymbol{H}^{d_{max}}); \operatorname{mean} ({\boldsymbol{H}}^{d_{max}}); \boldsymbol{g}^{d_{max}}]) \\
        & P(\widetilde{y} | \boldsymbol{v}) = \operatorname{softmax}(\boldsymbol{W}_{cls}\boldsymbol{v} + \boldsymbol{b}_{cls})
        \label{equ_cls_pred}
    \end{split}
\end{equation}
where $P(\widetilde{y} | \boldsymbol{v})$ is the probability distribution over the label set, and $\boldsymbol{W}_{cls}$ and $\boldsymbol{b}_{cls}$ are trainable parameters. Afterwards, the most probable label $\hat{y}$ is chosen from the above probability distribution drawn by Eq. (\ref{equ_cls_pred}), computed as:
\begin{equation}
\setlength{\abovedisplayskip}{3pt}
\setlength{\belowdisplayskip}{3pt}
    \begin{split}
        & \hat{y} = \mathop{\arg\max} P (\widetilde{y} | \boldsymbol{v} ) \\
    \end{split}
    \label{background_intent_pred}
\end{equation}

For training, we denote $y_i$ as golden label for the $i$-{th} sample, and $|S|$ as the size of the label set, then the loss function is computed as cross entropy:
\begin{equation}
\setlength{\abovedisplayskip}{3pt}
\setlength{\belowdisplayskip}{3pt}
    \begin{split}
      & loss = - \sum\limits_{i=1}^{|S|}{y_i} log(P_{i}(\widetilde{y} | \boldsymbol{v}))  \\
    \end{split}
\end{equation}

\section{Experiments}
\begin{table*}[t!]
\begin{center}
\scalebox{0.9}{
\begin{tabular}{l|c c c c c c c}
\hline
Data / Model & MS-Trans. & Transformer $\dagger$ & Star-Trans. $\dagger$ & 3L-BiLSTMs $\dagger$ & S-LSTM $\dagger$ & RCRN & Ours \\
\hline
Apparel      & 86.5 & 87.3  & 88.7 & 89.2 & 89.8 & 90.5 & \textbf{91.0} \\ 
Baby         & 86.3 & 85.6  & 88.0 & 88.5 & 89.3 & 89.0 & \textbf{89.8} \\
Books        & 87.8 & 85.3  & 86.9 & 87.2 & 88.8 & 88.0 & \textbf{89.0} \\
Camera       & 89.5 & 89.0  & 91.8 & 89.7 & 91.5 & 90.5 & \textbf{92.3} \\
Dvd          & 86.5 & 86.3  & 87.4 & 86.0 & 89.0 & 86.8 & \textbf{88.8} \\
Electronics  & 84.3 & 86.5  & 87.2 & 87.0 & 86.8 & \textbf{89.0} & 88.3 \\
Health       & 86.8 & 87.5  & 89.1 & 89.0 & 89.0 & 90.5 & \textbf{90.8} \\
Imdb         & 85.0 & 84.3  & 85.0 & 88.0 & 87.6 & \textbf{89.8} & 89.5 \\
Kitchen      & 85.8 & 85.5  & 86.0 & 84.5 & 86.6 & 86.0 & \textbf{88.5} \\
Magazines    & 91.8 & 91.5  & 91.8 & 92.5 & 93.3 & \textbf{94.8} & 94.3 \\
Mr           & 78.3 & 79.3  & 79.0 & 77.7 & 79.0 & 79.0 & \textbf{79.8} \\
Music        & 81.5 & 82.0  & 84.7 & 85.7 & 84.0 & 86.0 & \textbf{86.5} \\
Software     & 87.3 & 88.5  & 90.9 & 90.3 & 90.3 & 90.8 & \textbf{91.5} \\
Sports       & 85.5 & 85.8  & 86.8 & 86.5 & 86.0 & \textbf{88.0} & 87.0 \\
Toys         & 87.8 & 87.5  & 85.5 & 90.5 & 88.0 & 90.8 & \textbf{91.0} \\
Video        & 88.4 & 90.0  & 89.3 & 87.8 & 89.6 & 88.5 & \textbf{90.2} \\
\hline
Avg          & 86.2 & 86.4 & 87.4 & 87.5 & 88.0 & 88.6 & \textbf{89.3}  \\
\hline
 \end{tabular}}
\end{center} \vspace{-3pt}
\caption{Accuracy scores (\%) on the Amazon-16 datasets. $\dagger$ is our implementations with several recent advanced techniques ({\em e.g., label smoothing}) under the unified setting. We establish new state-of-the-art results on 12 of total 16 datasets, and outperform the existing highest average score (+0.7\%).
}   
 \vspace{-3pt}
\label{amazon16_result}
\end{table*}

\begin{table*}[t!]
\begin{center}
\scalebox{0.85}{
\begin{tabular}{l|c c c c c c c c |c}
\hline

{Models / Dataset}  & TREC  & MR & Subj   &  IMDB & AG. &  DBP.  & Yelp P. & Yelp F. & Avg.\\
\hline
RCRN \cite{RCRN_2018}                 & 96.20  & --     & --    & 92.80  & --    & --     & --      & -- & --     \\
Cove  \cite{Cove_2017}                & 95.80  & --     & --    & 91.80  & --    & --     & --      & -- & --     \\
Text-CNN \cite{textcnn_2014}          & 93.60  & 81.50  & 93.40 & --     & --    & --     & --      & -- & --     \\
Multi-QT \cite{MR_efficient_2018}     & 92.80  & 82.40  & 94.80 & --     & --    & --     & --      & -- & --     \\
AdaSent \cite{subj_self_2015}         & 92.40  & 83.10  & \textbf{95.50} & --     & --    & --     & --  & --    & --      \\
CNN-MCFA \cite{subj_trans_2018}       & 94.20  & 81.80  & 94.40 & --     & --    & --     & --      & --      & -- \\
Capsule-B \cite{capsule_2018}         & 92.80     & 82.30     & 93.80    & --     & 92.60       & -- & -- & --  & -- \\ 
DNC+CUW \cite{less_memory_2019}       & --     & --     & --    & --   &  93.90  & -- & 96.40 & 65.60 & -- \\
Region-Emb \cite{region_emb_2018}   & --     & --     & --    & --  & 92.80 & 98.90 & 96.40 & 64.90 & -- \\
Char-CNN \cite{char_cnn_2015}         & --     & --     & --    & --     & 90.49 & 98.45  & 95.12   & 62.05   & -- \\
DPCNN \cite{DPCNN_2017} & --     & --     & --    & --  & 93.13  & 99.12  &  \textbf{97.36} & 69.42 & -- \\
DRNN \cite{DRNN_2018}                 & --     & --     & --    & --     & 94.47 & \textbf{99.19}  & 97.27   & 69.15   & -- \\
SWEM-concat \cite{swem_2018} & 92.20 & 78.20 & 93.00 & -- & 92.66 & 98.57 & 95.81 & 63.79 & -- \\
\hline
Star-Transformer \cite{star_transformer_2019} $\dagger$         & 93.00  & 79.76  & 93.40 & 94.52  & 92.50 & 98.62  & 94.20   & 63.21 & 88.65  \\
Transformer \cite{Transformer_2017}  $\dagger$              & 92.00  & 80.75  & 94.00 & 94.58  & 93.66 & 98.27  & 95.07   & 63.40 & 88.97 \\
S-LSTM \cite{slstm_2018} $\dagger$    & 96.00  & 82.92  & 95.10 & 94.92  & 94.55 & 99.02  & 96.22   & 65.37  & 90.51 \\
3L-BiLSTMs \shortcite{LSTM_1997} $\dagger$ & 95.60  & \textbf{83.50}  & 95.30 & 93.89  & 93.99 & 98.97  & 96.86   & 66.86  & 90.62   \\
Ours                                  & \textbf{96.40}  & 83.42  & 95.50 & \textbf{96.27}  & \textbf{94.93} & 99.16  & 97.34  & \textbf{70.14}  & \textbf{91.64} \\
\hline
\end{tabular}}
\end{center} \vspace{-3pt}
\caption{Accuracy scores (\%) on modestly sized and large-scaled  datasets.  $\dagger$ is our implementations with several recent advanced techniques and analogous parameter sizes. Our model achieves new state-of-the art results on 4 of 8 datasets under the same settings.
}  \vspace{-3pt}
\label{cls_result}
\end{table*}

\subsection{Task and Datasets}
Text classification is a classic task for NLP, which aims to assign a predefined category to free-text documents \cite{char_cnn_2015}, and is generally evaluated by accuracy scores. Generally, The number of categories may range from two to more, which correspond to binary and fine-grained classification. We conduct extensive experiments on the 24 popular datasets collected from diverse domains ({\em e.g.,} \textit{sentiment}, \textit{question}), and range from modestly sized to large-scaled. The statistics of these datasets are listed in Table \ref{data_statistics}.

\subsection{Implementation Details}
We apply dropout \cite{dropout_2014} to word embeddings and hidden states with a rate of 0.3 and 0.2 respectively. Models are optimized by the Adam optimizer \cite{Adam_2014} with gradient clipping of 5 \cite{gradient_clip_2013}. 
The initial learning rate $\alpha$ is set to 0.001, and decays with the increment of training steps. For datasets without standard train/test split, we adopt 5-fold cross validation. For datasets without a development set, we randomly sample 10\% training samples as the development set \footnote{Since the size of train set in Amazon-16 ({\em i.e., 1600 }) is much smaller than others, our preliminary experiments show the random train/dev split has non-trivial effect on performance. Thus we give up development set, and re-implement baseline models under the same setting. } .
One layer CNN with a filter of size 3 and max pooling are utilized to generate 50d character-level word embeddings. The novel depth embedding is a trainable matrix in 50d. The cased 300d Glove is adapted to initialize word embeddings, and keeps fixed when training. We conduct hyper-parameters tuning to find the proper value of layer size $L$ (finally set to 9), and empirically set hidden size to 400 \footnote{We slightly adjust hidden size among different models to make sure analogous parameter sizes.}, temperature $\tau$ to 0.001.

\subsection{Main Results} 
Please note that current hot pre-trained language models ({\em e.g.,} BERT \cite{bert_2019}, XLNet \cite{xlnet_2019}) are not directly comparable with our work due to their huge additional corpora. We believe further improvements when utilizing these orthogonal works.

\paragraph{Results on Amazon-16.}
The results on 16 Amazon reviews are shown in Table \ref{amazon16_result}, where our model achieves state-of-the-art results on 12 datasets, and reports a new highest average score. 
The average score gains over 3-layer stacked Bi-LSTMs (+1.8\%), and the S-LSTM (+1.3\%) are also notable.
Strong baselines such as Star-Transformer \cite{star_transformer_2019} and Recurrently Controlled Recurrent Networks (RCRN) \cite{RCRN_2018} are also outperformed by our model.

\begin{table*}[t!]
\begin{center}
\scalebox{0.9}{
\begin{tabular}{c| l |c c| c c}
\hline
 \textbf{\#} & \textbf{Model} & Accuracy  & $\Delta_{acc}$ & Speed &  $\Delta_{speed}$ \\
\hline
0 & Ours  & \textbf{96.27 $\pm$ 0.13 } & -- & 57 $\pm$ 1.7 & -- \\
\hline
1 & w/o adaptive-depth mechanism & 96.10 $\pm$ 0.08 & -0.17  & 16 $\pm$ 0.3 & -41 \\
\hline
2 & w/o Bi-LSTMs  & 95.25 $\pm$ 0.25 & -1.02 & 49 $\pm$ 1.3  & -8 \\ 
3 & \ \ \ \ w/ sinusoidal position embedding & 95.61  $\pm$ 0.19 & -0.66 & \textbf{63 $\pm$ 1.1} & +6 \\ 
4 & \ \ \ \ w/ learned position embedding & 95.72 $\pm$ 0.15 & -0.55 & 60 $\pm$ 1.0 & +3 \\ 
\hline
-- & w/o Gumbel Max  & -- & -- & -- & -- \\
5 & \ \ \ \ w/ hard selection & 95.65 $\pm$ 0.39 & -0.62 & 57 $\pm$ 1.2 & 0 \\ 
6 & \ \ \ \ w/ soft selection  & 95.71 $\pm$ 0.57 & -0.56  & 57 $\pm$ 1.4 & 0 \\ 
\hline
\end{tabular}}
\end{center} 
\vspace{-3pt}
\caption{ 
Ablation experiments on IMDB test set. We run each model variant for three times and report the mean and standard deviation. `$\Delta_{acc}$' and `$\Delta_{speed}$' denote relative improvements of accuray and speed over `Ours' respectively.  {\em e.g.}, the `+6' in `$\Delta_{speed}$' denotes the variant processes 6 more samples than `Ours' per second.  
} 
\vspace{-5pt}
\label{ablation}
\end{table*}
 
\paragraph{Results on larger benchmarks.}
From the results on larger corpora listed in Table \ref{cls_result}, we also observe consistent and significant improvements over the conventional S-LSMT (+1.1\%) and other strong baseline models ({\em e.g.,} the transformer (+2.9\%), the star-transformer (+3.0\%)). More notably, the superiority of our model over baselines are more obvious with the growth of corpora size. Given  only training data and the ubiquitous word embeddings (Glove), our model achieves state-of-the-art performance on the TREC, IMDB, AGs News and Yelp Full datasets, and comparable results on other sets. 


\begin{figure}[t!]
\vspace*{-10pt}
\begin{center}
     \scalebox{0.45}{
       \includegraphics[width=1\textwidth]{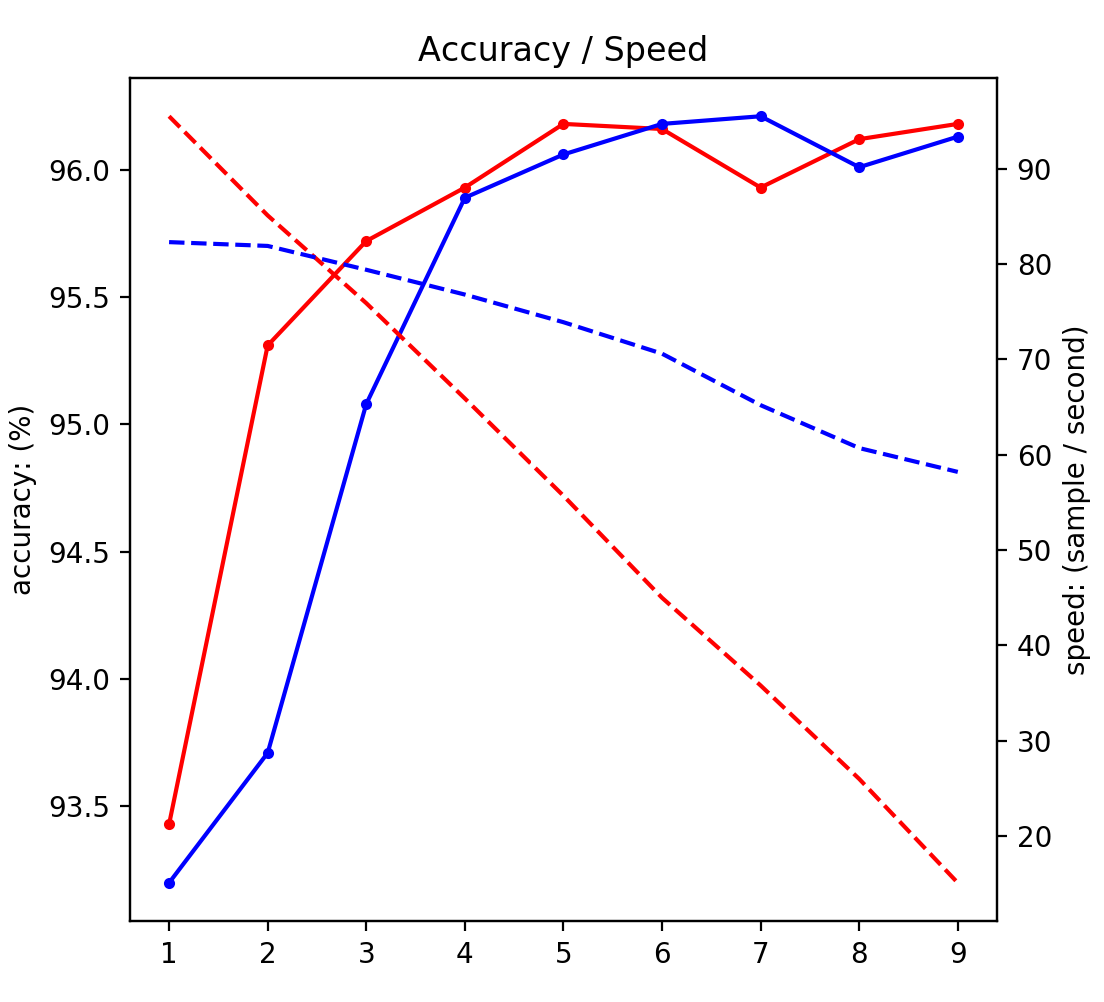}
      }\vspace{-10pt}
      \caption{Accuracy and speed \protect\footnotemark\, for full-depth (red lines) and adaptive-depth (blue lines) models on the IMDB test set, where X-axis refer to the maximum of layer $L$, and accuracy/speed are drawn by solid/dashed lines, respectively. } \label{vs_full_layer}
      \vspace{-10pt}
 \end{center}
\end{figure}
\footnotetext{Number of samples calculated in one second on one Tesla P40 GPU with the batch size of 100.}  

\begin{figure*}[t!]
\begin{center}
     \scalebox{0.9}{
      \includegraphics[width=1\textwidth]{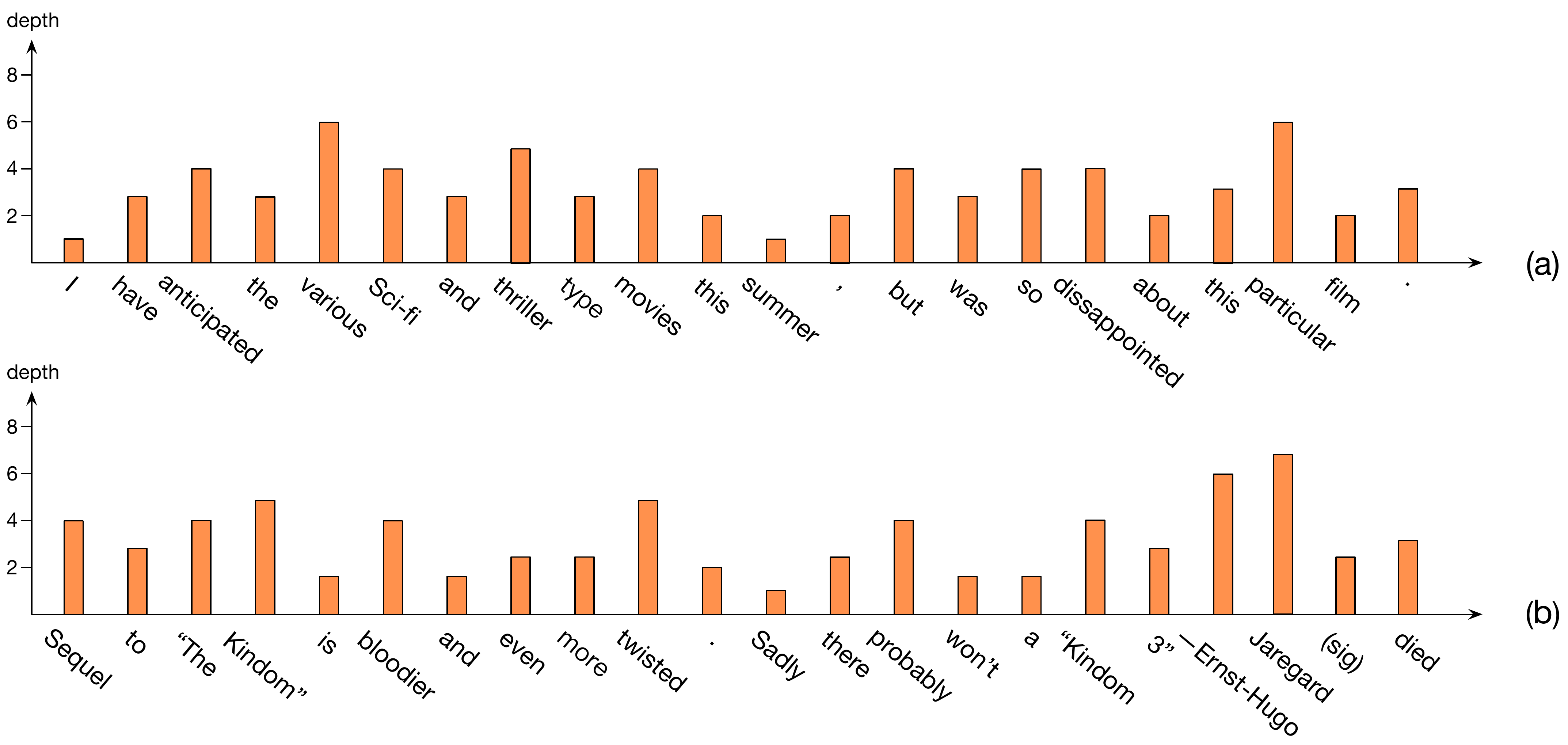}
      }\vspace{-10pt}
      \caption{Histogram of \textit{executed} depths over words of two random examples in the IMDB sentiment dataset with negative (a) and  positive (b) labels, respectively. } \label{case_study}  \vspace{-15pt}
 \end{center}
\end{figure*}

\section{Analysis}
We conduct analytical experiments on a modestly sized dataset ({\em i.e.,} IMDB) to offer more insights and elucidate the properties of our model.

\subsection{Compared with Full-depth Model}
\label{speed_acc_section}
In our model, the depth is dynamically varying at each word position, and thus it is intuitive to compare the performance with a full-depth model in both terms of accuracy and speed.
For fair comparisons, we conduct two groups of experiments on the IMDB test set only with difference in using adaptive-depth mechanism or not.
As shown in Figure \ref{vs_full_layer}, when $L \in [1,4]$, the full-depth model consistently outperforms our depth-adaptive model, due to the insufficient modeling in the lower layers. We also observe the accuracy gap gradually decreasing with the growth of layer number. As $L \in [5,9]$, both models perform nearly identical results, but the evident superiority appears when we focus on the speed. Concretely, the speed of full-depth model decays almost linearly with the increase of depths. Howerver, our depth-adaptive model shows a more flat decrease against the increase of depths. Specifically, at the $9$-{th} layer, our model performs 3$\times$ faster than the full-depth model, which amounts to the speed of a full-depth model with $5$ layers, namely only half parameters.

\subsection{Ablation Experiments}
\label{ablation_section}
We conduct ablation experiments to investigate the impacts of our depth-adaptive mechanism, and different strategies of depth selection and how to inject sequential information.

As listed in Table \ref{ablation}, the adaptive depth mechanism has a slight influence on performance, but is of great matter in terms of speed (row 1 {\em vs.} row 0), which is consistent with our observations in Section \ref{speed_acc_section}.

Results in terms of injecting sequential information is shown from row $2$ to $4$ in Table \ref{ablation}. Although the additional Bi-LSTMs layer decreases the speed to some extend, its great effect on accuracy indicates this recurrent inductive bias is necessary and effective for text representation.
Two position embedding alternatives (row $3$ and $4$) could also alleviate the lack of sequential information to a certain extent and meanwhile get rid of the time-inefficient problem of RNN (row $2$).

In respect of depth selection (row $5$ and $6$), the Gumbel-Max technique provides a more robust depth estimation, compared with direct (hard or soft) selections.

\subsection{Case Study}
We choose two examples from the IMDB train set with positive and negative labels, and their depth distributions are shown in Figure \ref{case_study}. 
Our model successfully pays more attentions to words ({\em e.g.,} 'thriller', 'twisted') that are relatively more difficult to learn, and allocates fewer computation steps for common words ({\em e.g.,} 'film', 'and'). 

\section{Related Work}

\paragraph{Extensions of the S-LSTM.} \citet{star_transformer_2019}
 enhance neural gates in the S-LSTM with self-attention mechanism \cite{Transformer_2017}, and propose the Star-Transformer, which has shown promising performance for sentence modeling. 
\citet{slstm_re_2018} extend the conventional S-LSTM to the graph state LSTM  for $N$-ary Relation Extraction. Inspired by the rich nodes communications in the S-LSTM, \citet{densely_slstm_2019} propose the extend Levi graph with a global node.
Different from these work, we mainly focus on the problem of computational efficiency in the S-LSTM, and thus propose a depth-adaptive mechanism. Extensive experiments suggest our method achieves a good accuracy-speed trade off. 

\paragraph{Conditional Computation.}
Our work is inspired by conditional computation, where only parts of the network are selectively activated according to gating units \cite{condition_2013} or a learned policy \cite{conditional_rl_2015}. 
A related architecture, known as Adaptive Computation Time (ACT) \cite{ACT_2016}, employs a halting unit upon each word when sequentially reading a sentence.
The halting unit determines the probability that computation should continue or stop step-by-step. 
ACT has been extended to control the layers of the Residual Networks \cite{ACT_resnet_2017} and the Universal Transformer \cite{UT_2019}. Unlike the continuous layer-wise prediction to determine a stop probability in the ACT, we provide an effective alternative method with more straightforward modeling, which directly predicts the depth distribution among words simultaneously. 
Another concurrent work named `Depth-Adaptive Transformer' \cite{depth_ada_transformer_2020} proposes to dynamically reduce computational burdens for the decoder in the sequence-to-sequence framework. 
In this paper, we investigate different ways to obtain the depths ({\em e.g.,} Gumbel-Max), and propose a novel depth embedding to endow the model with depth-specific view.
Another group of work explores to conduct conditional computation inside the dimension of  neural network representations\cite{ACT_dimensition_2017,ON_LSTM_2019}, instead of activating partial layers of model, {\em e.g.}, adaptive depths in our method.

\section{Conclusion}
We propose a depth-adaptive mechanism to allow the model itself to `ponder' and `determine' the number of depths for different words. In addition, we investigate different approaches to inject sequential information into the S-LSTM. Empirically, our model brings consistent improvements in terms of both accuracy and speed over the conventional S-LSTM, and achieves state-of-the-art results on 16 out of 24 datasets. In the future, we would like to extend our model on some generation tasks, {\em e.g.}, machine translation, and investigate how to introduce explicit supervision for the depth distribution.

\section*{Acknowledgments}
Liu, Chen and Xu are supported by the National Natural Science Foundation of China (Contract 61370130, 61976015, 61473294 and 61876198), and the Beijing Municipal Natural Science Foundation (Contract 4172047), and the International Science and Technology Cooperation Program of the Ministry of Science and Technology (K11F100010). We sincerely thank the anonymous reviewers for their thorough reviewing and valuable suggestions.

\bibliography{acl2020}

\begin{thebibliography}{46}
\expandafter\ifx\csname natexlab\endcsname\relax\def\natexlab#1{#1}\fi

\bibitem[{Amplayo et~al.(2018)Amplayo, Lee, Yeo, and Hwang}]{subj_trans_2018}
Reinald~Kim Amplayo, Kyungjae Lee, Jinyeong Yeo, and Seung-won Hwang. 2018.
\newblock Translations as additional contexts for sentence classification.
\newblock In \emph{Proceedings of the 27th International Joint Conference on
  Artificial Intelligence}.

\bibitem[{Beck et~al.(2018)Beck, Haffari, and Cohn}]{slstm_mt_2018}
Daniel Beck, Gholamreza Haffari, and Trevor Cohn. 2018.
\newblock Graph-to-sequence learning using gated graph neural networks.
\newblock In \emph{Proceedings of the 56th Annual Meeting of the Association
  for Computational Linguistics}.

\bibitem[{Bengio et~al.(2015)Bengio, Bacon, Pineau, and
  Precup}]{conditional_rl_2015}
Emmanuel Bengio, Pierre-Luc Bacon, Joelle Pineau, and Doina Precup. 2015.
\newblock Conditional computation in neural networks for faster models.
\newblock \emph{arXiv}.

\bibitem[{Bengio et~al.(2013)Bengio, Léonard, and Courville}]{condition_2013}
Yoshua Bengio, Nicholas Léonard, and Aaron Courville. 2013.
\newblock Estimating or propagating gradients through stochastic neurons for
  conditional computation.
\newblock \emph{arXiv}.

\bibitem[{Chen et~al.(2018)Chen, Firat, Bapna, Johnson, Macherey, Foster,
  Jones, Schuster, Shazeer, Parmar, and et~al.}]{best_2018}
Mia~Xu Chen, Orhan Firat, Ankur Bapna, Melvin Johnson, Wolfgang Macherey,
  George Foster, Llion Jones, Mike Schuster, Noam Shazeer, Niki Parmar, and
  et~al. 2018.
\newblock The best of both worlds: Combining recent advances in neural machine
  translation.
\newblock In \emph{Proceedings of the 56th Annual Meeting of the Association
  for Computational Linguistics}.

\bibitem[{Dehghani et~al.(2019)Dehghani, Gouws, Vinyals, Uszkoreit, and Łukasz
  Kaiser}]{UT_2019}
Mostafa Dehghani, Stephan Gouws, Oriol Vinyals, Jakob Uszkoreit, and Łukasz
  Kaiser. 2019.
\newblock Universal transformers.
\newblock In \emph{Proceedings of the Seventh International Conference on
  Learning Representations}.

\bibitem[{Devlin et~al.(2019)Devlin, Chang, Lee, and Toutanova}]{bert_2019}
Jacob Devlin, Ming-Wei Chang, Kenton Lee, and Kristina Toutanova. 2019.
\newblock {BERT}: Pre-training of deep bidirectional transformers for language
  understanding.
\newblock In \emph{Proceedings of the 2019 Conference of the North {A}merican
  Chapter of the Association for Computational Linguistics: Human Language
  Technologies, Volume 1 (Long and Short Papers)}. Association for
  Computational Linguistics.

\bibitem[{Elbayad et~al.(2019)Elbayad, Gu, Grave, and
  Auli}]{depth_ada_transformer_2020}
Maha Elbayad, Jiatao Gu, Edouard Grave, and Michael Auli. 2019.
\newblock Depth-adaptive transformer.
\newblock \emph{arXiv preprint arXiv:1910.10073}.

\bibitem[{Figurnov et~al.(2017)Figurnov, Collins, Zhu, Zhang, Huang, Vetrov,
  and Salakhutdinov}]{ACT_resnet_2017}
Michael Figurnov, Maxwell~D. Collins, Yukun Zhu, Li~Zhang, Jonathan Huang,
  Dmitry Vetrov, and Ruslan Salakhutdinov. 2017.
\newblock Spatially adaptive computation time for residual networks.
\newblock In \emph{Proceedings of the IEEE Conference on Computer Vision and
  Pattern Recognition}.

\bibitem[{Gehring et~al.(2017)Gehring, Auli, Grangier, Yarats, and
  Dauphin}]{pos_emb_2017}
Jonas Gehring, Michael Auli, David Grangier, Denis Yarats, and Yann~N Dauphin.
  2017.
\newblock Convolutional sequence to sequence learning.
\newblock In \emph{Proceedings of the 34th International Conference on Machine
  Learning-Volume 70}, pages 1243--1252. JMLR. org.

\bibitem[{Graves(2016)}]{ACT_2016}
Alex Graves. 2016.
\newblock Adaptive computation time for recurrent neural networks.
\newblock \emph{arXiv}.

\bibitem[{Gumbel(1954)}]{gumbel_1954}
Emil~Julius Gumbel. 1954.
\newblock Statistical theory of extreme values and some practical applications.
\newblock \emph{NBS Applied Mathematics Series}.

\bibitem[{Guo et~al.(2019{\natexlab{a}})Guo, Qiu, Liu, Shao, Xue, and
  Zhang}]{star_transformer_2019}
Qipeng Guo, Xipeng Qiu, Pengfei Liu, Yunfan Shao, Xiangyang Xue, and Zheng
  Zhang. 2019{\natexlab{a}}.
\newblock Star-transformer.
\newblock In \emph{Proceedings of the 2019 Conference of the North {A}merican
  Chapter of the Association for Computational Linguistics: Human Language
  Technologies, Volume 1 (Long and Short Papers)}.

\bibitem[{Guo et~al.(2019{\natexlab{b}})Guo, Zhang, Teng, and
  Lu}]{densely_slstm_2019}
Zhijiang Guo, Yan Zhang, Zhiyang Teng, and Wei Lu. 2019{\natexlab{b}}.
\newblock Densely connected graph convolutional networks for graph-to-sequence
  learning.
\newblock \emph{Transactions of the Association for Computational Linguistics},
  7:297--312.

\bibitem[{Hochreiter and Schmidhuber(1997)}]{LSTM_1997}
Sepp Hochreiter and Jürgen Schmidhuber. 1997.
\newblock \href {https://doi.org/10.1162/neco.1997.9.8.1735} {Long short-term
  memory}.
\newblock \emph{Neural Computation}, 9(8):1735--1780.

\bibitem[{Jernite et~al.(2017)Jernite, Grave, Joulin, and
  Mikolov}]{ACT_dimensition_2017}
Yacine Jernite, Edouard Grave, Armand Joulin, and Tomas Mikolov. 2017.
\newblock Variable computation in recurrent neural networks.
\newblock In \emph{Proceedings of the Fifth International Conference on
  Learning Representations}.

\bibitem[{Johnson and Zhang(2017)}]{DPCNN_2017}
Rie Johnson and Tong Zhang. 2017.
\newblock Deep pyramid convolutional neural networks for text categorization.
\newblock In \emph{Proceedings of the 55th Annual Meeting of the Association
  for Computational Linguistics (Volume 1: Long Papers)}. Association for
  Computational Linguistics.

\bibitem[{Kim(2014)}]{textcnn_2014}
Yoon Kim. 2014.
\newblock Convolutional neural networks for sentence classification.
\newblock In \emph{Proceedings of the 2014 Conference on Empirical Methods in
  Natural Language Processing}.

\bibitem[{Kingma and Ba(2014)}]{Adam_2014}
Diederik~P Kingma and Jimmy Ba. 2014.
\newblock Adam: A method for stochastic optimization.
\newblock \emph{arXiv}.

\bibitem[{Le et~al.(2019)Le, Tran, and Venkatesh}]{less_memory_2019}
Hung Le, Truyen Tran, and Svetha Venkatesh. 2019.
\newblock Learning to remember more with less memorization.
\newblock In \emph{Proceedings of the Seventh International Conference on
  Learning Representations}.

\bibitem[{Li and Roth(2002)}]{trec_2002}
Xin Li and Dan Roth. 2002.
\newblock Learning question classifiers.
\newblock In \emph{Proceedings of the 19th international conference on
  Computational linguistics-Volume 1}, pages 1--7. Association for
  Computational Linguistics.

\bibitem[{Liu et~al.(2017)Liu, Qiu, and Huang}]{16_cls_data_17}
Pengfei Liu, Xipeng Qiu, and Xuanjing Huang. 2017.
\newblock Adversarial multi-task learning for text classification.
\newblock In \emph{Proceedings of the 55th Annual Meeting of the Association
  for Computational Linguistics}.

\bibitem[{Liu et~al.(2019)Liu, Meng, Zhang, Zhou, Chen, and Xu}]{cmnet_2019}
Yijin Liu, Fandong Meng, Jinchao Zhang, Jie Zhou, Yufeng Chen, and Jinan Xu.
  2019.
\newblock \href {https://www.aclweb.org/anthology/D19-1097} {{CM}-net: A novel
  collaborative memory network for spoken language understanding}.
\newblock In \emph{Proceedings of the 2019 Conference on Empirical Methods in
  Natural Language Processing and the 9th International Joint Conference on
  Natural Language Processing (EMNLP-IJCNLP)}, Hong Kong, China. Association
  for Computational Linguistics.

\bibitem[{Logeswaran and Lee(2018)}]{MR_efficient_2018}
Lajanugen Logeswaran and Honglak Lee. 2018.
\newblock An efficient framework for learning sentence representations.
\newblock \emph{arXiv}.

\bibitem[{Maas et~al.(2011)Maas, Daly, Pham, Huang, Ng, and Potts}]{IMDB_2011}
Andrew~L Maas, Raymond~E Daly, Peter~T Pham, Dan Huang, Andrew~Y Ng, and
  Christopher Potts. 2011.
\newblock Learning word vectors for sentiment analysis.
\newblock In \emph{Proceedings of the 49th Annual Meeting of the Association
  for Computational Linguistics: Human language technologies-volume 1}, pages
  142--150. Association for Computational Linguistics.

\bibitem[{Maddison et~al.(2014)Maddison, Tarlow, and Minka}]{gumbel_2014}
Chris~J Maddison, Daniel Tarlow, and Tom Minka. 2014.
\newblock A* sampling.
\newblock In \emph{Advances in Neural Information Processing Systems}.

\bibitem[{McCann et~al.(2017)McCann, Bradbury, Xiong, and Socher}]{Cove_2017}
Bryan McCann, James Bradbury, Caiming Xiong, and Richard Socher. 2017.
\newblock Learned in translation: Contextualized word vectors.
\newblock In \emph{Advances in Neural Information Processing Systems}.

\bibitem[{Pang and Lee(2004)}]{subj_2004}
Bo~Pang and Lillian Lee. 2004.
\newblock A sentimental education: Sentiment analysis using subjectivity
  summarization based on minimum cuts.
\newblock In \emph{Proceedings of the 42nd Annual Meeting of the Association
  for Computational Linguistics}.

\bibitem[{Pang and Lee(2005)}]{MR_2005}
Bo~Pang and Lillian Lee. 2005.
\newblock Seeing stars: Exploiting class relationships for sentiment
  categorization with respect to rating scales.
\newblock In \emph{Proceedings of the 43rd Annual Meeting of the Association
  for Computational Linguistics}.

\bibitem[{Pascanu et~al.(2013)Pascanu, Mikolov, and
  Bengio}]{gradient_clip_2013}
Razvan Pascanu, Tomas Mikolov, and Yoshua Bengio. 2013.
\newblock On the difficulty of training recurrent neural networks.
\newblock \emph{The journal of machine learning research}, pages 1310--1318.

\bibitem[{Qiao et~al.(2018)Qiao, Huang, Niu, Li, Dong, He, Yu, and
  Wu}]{region_emb_2018}
Chao Qiao, Bo~Huang, Guocheng Niu, Daren Li, Daxiang Dong, Wei He, Dianhai Yu,
  and Hua Wu. 2018.
\newblock A new method of region embedding for text classification.
\newblock In \emph{Proceedings of the Sixth International Conference on
  Learning Representations}.

\bibitem[{Santos and Zadrozny(2014)}]{first_CNN_2014}
C{\'i}cero Nogueira~Dos Santos and Bianca Zadrozny. 2014.
\newblock Learning character-level representations for part-of-speech tagging.
\newblock In \emph{Proceedings of the 31th international conference on
  Computational linguistics}.

\bibitem[{Shen et~al.(2018)Shen, Wang, Wang, Min, Su, Zhang, Li, Henao, and
  Carin}]{swem_2018}
Dinghan Shen, Guoyin Wang, Wenlin Wang, Martin~Renqiang Min, Qinliang Su, Yizhe
  Zhang, Chunyuan Li, Ricardo Henao, and Lawrence Carin. 2018.
\newblock Baseline needs more love: On simple word-embedding-based models and
  associated pooling mechanisms.
\newblock In \emph{Proceedings of the 56th Annual Meeting of the Association
  for Computational Linguistics}.

\bibitem[{Shen et~al.(2019)Shen, Tan, Sordoni, and Courville}]{ON_LSTM_2019}
Yikang Shen, Shawn Tan, Alessandro Sordoni, and Aaron Courville. 2019.
\newblock Ordered neurons: Integrating tree structures into recurrent neural
  networks.
\newblock In \emph{Proceedings of the Seventh International Conference on
  Learning Representations}.

\bibitem[{Song et~al.(2018)Song, Zhang, Wang, and Gildea}]{slstm_re_2018}
Linfeng Song, Yue Zhang, Zhiguo Wang, and Daniel Gildea. 2018.
\newblock N-ary relation extraction using graph-state lstm.
\newblock In \emph{Proceedings of the 2018 Conference on Empirical Methods in
  Natural Language Processing}.

\bibitem[{Srivastava et~al.(2014)Srivastava, Hinton, Krizhevsky, Sutskever, and
  Salakhutdinov}]{dropout_2014}
Nitish Srivastava, Geoffrey Hinton, Alex Krizhevsky, Ilya Sutskever, and Ruslan
  Salakhutdinov. 2014.
\newblock Dropout: A simple way to prevent neural networks from overfitting.
\newblock \emph{The journal of machine learning research}, 15(1):1929--1958.

\bibitem[{Tay et~al.(2018)Tay, Tuan, and Hui}]{RCRN_2018}
Yi~Tay, Luu~Anh Tuan, and Siu~Cheung Hui. 2018.
\newblock Recurrently controlled recurrent networks.
\newblock In \emph{Advances in Neural Information Processing Systems}.

\bibitem[{Vaswani et~al.(2017)Vaswani, Shazeer, Parmar, Uszkoreit, Jones,
  Gomez, Kaiser, and Polosukhin}]{Transformer_2017}
Ashish Vaswani, Noam Shazeer, Niki Parmar, Jakob Uszkoreit, Llion Jones,
  Aidan~N Gomez, {\L}ukasz Kaiser, and Illia Polosukhin. 2017.
\newblock Attention is all you need.
\newblock In \emph{Advances in Neural Information Processing Systems}.

\bibitem[{Wang(2018)}]{DRNN_2018}
Baoxin Wang. 2018.
\newblock Disconnected recurrent neural networks for text categorization.
\newblock In \emph{Proceedings of the 56th Annual Meeting of the Association
  for Computational Linguistics}.

\bibitem[{Wang et~al.(2019)Wang, Ma, Liu, and Tang}]{r_transformer_2019}
Zhiwei Wang, Yao Ma, Zitao Liu, and Jiliang Tang. 2019.
\newblock R-transformer: Recurrent neural network enhanced transformer.
\newblock \emph{arXiv preprint arXiv:1907.05572}.

\bibitem[{Yang et~al.(2018)Yang, Zhao, Ye, Lei, Zhao, and Zhang}]{capsule_2018}
Min Yang, Wei Zhao, Jianbo Ye, Zeyang Lei, Zhou Zhao, and Soufei Zhang. 2018.
\newblock Investigating capsule networks with dynamic routing for text
  classification.
\newblock In \emph{Proceedings of the 2018 Conference on Empirical Methods in
  Natural Language Processing}, Brussels, Belgium. Association for
  Computational Linguistics.

\bibitem[{Yang et~al.(2019)Yang, Dai, Yang, Carbonell, Salakhutdinov, and
  Le}]{xlnet_2019}
Zhilin Yang, Zihang Dai, Yiming Yang, Jaime Carbonell, Ruslan Salakhutdinov,
  and Quoc~V Le. 2019.
\newblock Xlnet: Generalized autoregressive pretraining for language
  understanding.
\newblock \emph{arXiv preprint arXiv:1906.08237}.

\bibitem[{Yin et~al.(2019)Yin, Song, Su, Zeng, Zhou, and
  Luo}]{sentence_ordering_2019}
Yongjing Yin, Linfeng Song, Jinsong Su, Jiali Zeng, Chulun Zhou, and Jiebo Luo.
  2019.
\newblock Graph-based neural sentence ordering.
\newblock In \emph{Proceedings of the 28th International Joint Conference on
  Artificial Intelligence}.

\bibitem[{Zhang et~al.(2015)Zhang, Zhao, and LeCun}]{char_cnn_2015}
Xiang Zhang, Junbo Zhao, and Yann LeCun. 2015.
\newblock Character-level convolutional networks for text classification.
\newblock In \emph{Advances in Neural Information Processing Systems}, pages
  649--657.

\bibitem[{Zhang et~al.(2018)Zhang, Liu, and Song}]{slstm_2018}
Yue Zhang, Qi~Liu, and Linfeng Song. 2018.
\newblock Sentence-state {LSTM} for text representation.
\newblock In \emph{Proceedings of the 56th Annual Meeting of the Association
  for Computational Linguistics (Volume 1: Long Papers)}.

\bibitem[{Zhao et~al.(2015)Zhao, Lu, and Poupart}]{subj_self_2015}
Han Zhao, Zhengdong Lu, and Pascal Poupart. 2015.
\newblock Self-adaptive hierarchical sentence model.
\newblock In \emph{Proceedings of the 24th International Joint Conference on
  Artificial Intelligence}.

\end{thebibliography}
\bibliographystyle{acl_natbib}
\end{document}